\documentclass[11pt]{article}

\usepackage[final]{acl}
\usepackage{booktabs}
\usepackage{siunitx}
\usepackage{times}
\usepackage{latexsym}
\usepackage{booktabs}
\usepackage{graphicx}
\usepackage{float}  

\usepackage{pdfpages}

\usepackage[T1]{fontenc}

\usepackage[utf8]{inputenc}
\usepackage[ethiop,main=english]{babel}
\usepackage{microtype}

\usepackage{inconsolata}

\usepackage{graphicx}
\usepackage[most]{tcolorbox}
\usepackage{xcolor}

\usepackage{xspace}

\newcommand\blfootnote[1]{%
  \begingroup
  \renewcommand\thefootnote{}\footnote{#1}%
  \addtocounter{footnote}{-1}%
  \endgroup
}

\newcommand*{\gpt}{Gpt-4.1\xspace}
\newcommand*{\gemmathree}{\textit{Gemma-3-27b-it}}

\newcommand*{\llamaeight}{\textit{Llama 3.1 8B Instruct}}
\newcommand*{\llamaone}{\textit{Llama 3.2 1B Instruct}}

\newcommand*{\aya}{\textit{Aya 23 35B}}

\newcommand*{\oureng}{\textit{SFT Llama 3.1 8B (Eng)}}
\newcommand*{\ouramh}{\textit{SFT Llama 3.1 8B (Amh)}}
\newcommand*{\ourall}{\textit{SFT Llama 3.1 8B (All)}}

\title{\textit{AmharicStoryQA}: A Multicultural Story Question Answering Benchmark in Amharic}

\author{\normalsize Israel Abebe Azime$^{1,\ast }$,  Abenezer Kebede Angamo$^{2, \ast}$, Hana Mekonen Tamiru$^{3,\ast}$,   \\ 
\textbf{\normalsize Dagnachew Mekonnen Marilign$^{4}$, Philipp Slusallek$^{1}$, Seid Muhie Yimam$^{5}$, } \\
\textbf{\normalsize Dietrich Klakow$^{1}$ }  \\\\
\footnotesize
$^1$ Saarland University, $^2$ AIMS AMMI,  $^3$  Resonance AI4D Lab, Addis Ababa University,   
\\ 
 \footnotesize
     $^4$ HILCOE School of Computer Science and Technology, $^{5}$University of Hamburg
\\}

\begin{document}

\maketitle
\blfootnote{$^\ast$ Equal Contribution.}
\begin{abstract}
With the growing emphasis on multilingual and cultural evaluation benchmarks for large language models, language and culture are often treated as synonymous, and performance is commonly used as a proxy for a model’s understanding of a given language. In this work, we argue that such evaluations overlook meaningful cultural variation that exists within a single language. We address this gap by focusing on narratives from different regions of Ethiopia and demonstrate that, despite shared linguistic characteristics, region-specific and domain-specific content substantially influences language evaluation outcomes. To this end, we introduce \textbf{\textit{AmharicStoryQA}}, a long-sequence story question answering benchmark grounded in culturally diverse narratives from Amharic-speaking regions. Using this benchmark, we reveal a significant narrative understanding gap in existing LLMs, highlight pronounced regional differences in evaluation results, and show that supervised fine-tuning yields uneven improvements across regions and evaluation settings. Our findings emphasize the need for culturally grounded benchmarks that go beyond language-level evaluation to more accurately assess and improve narrative understanding in low-resource languages.
\end{abstract}

\section{Introduction}

Evaluating Large Language Models (LLMs) on low-resource languages is essential for uncovering their true language understanding capabilities in specific linguistic and cultural contexts, especially given that these models are predominantly trained on large-scale English corpora.

Despite increasing efforts to include more languages in LLM evaluations, particularly African-centric languages, there remains a notable gap in the creation of multicultural evaluation datasets that capture the diversity of linguistic and cultural contexts~\cite{alabi-etal-2025-charting}. Another important limitation in low-resource LLM evaluation is the scarcity of generation-focused datasets designed to simulate real-life model usage, which are essential for ensuring that benchmarking accurately reflects how LLMs perform in practical, everyday scenarios.

To capture the diversity of linguistic and cultural contexts, most recent works~\cite{havaldar-etal-2023-multilingual,azime-etal-2025-proverbeval} expand evaluations by adding more languages and incorporating culturally grounded information drawn from the regions they represent. This idea, \textit{which treats each language as being associated with a single cultural identity, has a solid foundation but still fails to capture the cultural variations that exist within the same language community}. These variations arise from differences in governmental structures, regional histories, social norms, migration patterns, and intra-country ethnic diversity. As a result, evaluations risk oversimplifying cultural nuance and underrepresenting the lived experiences of multilingual and multicultural societies

In this work, we introduce \textbf{\textit{AmharicStoryQA}}, a multicultural story-based question answering benchmark available in Amharic and English. The benchmark is constructed from pre-translated stories and folktales collected from nine different regions of Ethiopia. Our goal is to provide an evaluation resource that captures Ethiopia’s rich regional diversity while enabling the study of long-sequence question answering. \textit{AmharicStoryQA} focuses on culturally grounded narrative understanding and offers one of the first long-context QA benchmarks centered on Ethiopian stories across nine regions.

Based on the benchmark we introduce, we aim to answer the following research questions: (\textbf{RQ1}) \textit{How well do Large Language Models comprehend Amharic narratives?} This is measured through performance on a story-based question answering datasets. (\textbf{RQ2}) \textit{How cultural differences influence LLM performance in a single language? } and (\textbf{RQ3}) \textit{Can we improve LLM performance using the culturally grounded training dataset we introduce, demonstrating the importance of such data for addressing the limitations observed?} To support these questions, we further conduct human error analysis to better identify the linguistic and cultural factors contributing to model failures.

To address this question, our work makes the following contributions.

\begin{itemize}
    \item We introduce \textbf{\textit{AmharicStoryQA}}, a multicultural story-based question answering benchmark in Amharic and English, consisting of 571 training and 649 test examples derived from 244 stories collected across nine Ethiopian regions. The benchmark includes both multiple-choice question answering (MCQA) and open-ended generation questions, enabling comprehensive evaluation of model understanding across formats.~\footnote{https://hf.co/collections/israel/amharicstoryqa}

    \item We explore the zero-shot performance of seven opensource LLMs and analyze how their results vary across regions, languages, prompt languages, and task types with human error analysis.

    \item We explore the effect of supervised fine-tuning on improving the performance of selected LLMs.
\end{itemize}

\section{Related Work}

\paragraph{Low Resource LLM Evaluations.}
Low-resource conditions, characterized by the lack of datasets for specific tasks and domains, limited digitization of languages, and insufficient data to effectively train and improve models, are common across many African languages~\cite{nigatu-etal-2024-zenos}.
Evaluating Large Language Models (LLMs) in such settings presents unique challenges, primarily due to data scarcity and the dominance of English-centric benchmarks. Recent efforts have begun to address this gap by introducing multilingual evaluation benchmarks that include African languages~\cite{adelani-etal-2025-irokobench, ojo-etal-2025-afrobench}. Despite the great deal of work done to include number of tasks and languages and cultural values~\cite{romero2024cvqa} there is still a need to deeper exploration in single language multiple cultural evaluations.

\paragraph{Long Sequence LLM Evaluations.}
While the capability of LLMs to process long contexts is a rapidly evolving area of research, benchmarks evaluating this capability are predominantly available only for high-resource languages.
Recent works such as~\citet{alabi-etal-2025-afridoc} evaluate long-context document translation across five low-resource languages, including Amharic. Despite these efforts, there remains a critical absence of long-sequence benchmarks for Amharic that evaluate an LLM’s ability to maintain coherence and track narrative arcs over extended contexts, leaving a significant blind spot in understanding how modern LLMs handle extensive texts in Ethiopian languages.

\paragraph{Story Question Answering.}
Story-based question answering assesses the ability of the model to comprehend narrative structures, character motivations, and causal chains. 
AfriQA~\cite{ogundepo-etal-2023-cross} introduces a cross-lingual open-retrieval QA dataset covering seven African languages, evaluating models on their ability to retrieve and answer questions using evidence that is often available only in English or French. Similarly, Belebele~\cite{bandarkar-etal-2024-belebele} offers a parallel reading comprehension benchmark across 122 language variants, including Amharic. However, these evaluations typically rely on short passages or translated content, which may not fully capture the linguistic depth required for native-level understanding. Furthermore, multilingual benchmarks often treat languages such as Amharic as monolithic, overlooking the need for region-specific evaluations that account for local variations in dialect and usage.




\begin{figure*}
\centering
\includegraphics[width=\textwidth]{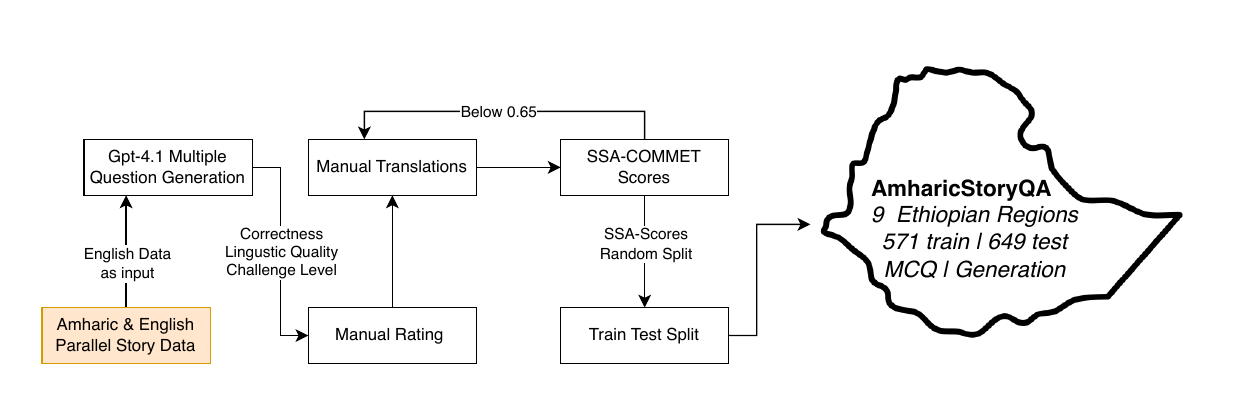}
\caption{\textbf{Overview of the \textit{AmharicStoryQA} dataset} creation process, from Amharic and English parallel stories through question generation, manual translation rating, quality control, and train–test split verification.}
\label{fig:AmharistoryQA}
\end{figure*}

\paragraph{Cultural Evaluations}
Truly robust LLMs must capture not only the syntactic structure of a language but also its cultural context and dialectal variation. Recent work has examined the fairness and robustness of LLMs across different dialects of English, even when a single language is used as the communication medium~\cite{srirag-etal-2025-evaluating}. Other studies leverage culturally grounded proverbs and sayings across multiple languages to more rigorously evaluate language understanding capabilities~\cite{liu-etal-2024-multilingual, azime-etal-2025-proverbeval}. In this work, we adopt a single-language, multi-cultural evaluation framework and extend prior efforts by introducing long-sequence, culturally grounded, story-based question answering, enabling a more comprehensive assessment of language understanding across cultural variations.

\paragraph{Amharic Question Answering}

Recent work such as~\cite{taffa-etal-2024-low,destaw-etal-2022-question} introduces an extensive Amharic question answering benchmark used to evaluate Amharic QA capabilities; however, it focuses primarily on open-ended question answering, which is more challenging to evaluate reliably. In this work, we contribute a multi-task story-based question answering dataset that spans multiple cultures in same language setting, resulting in broader cultural diversity and improved regional coverage additional to narrative style long sequence source documents that was not included in previous works.

\begin{figure*}
\centering
\includegraphics[width=0.7\textwidth]{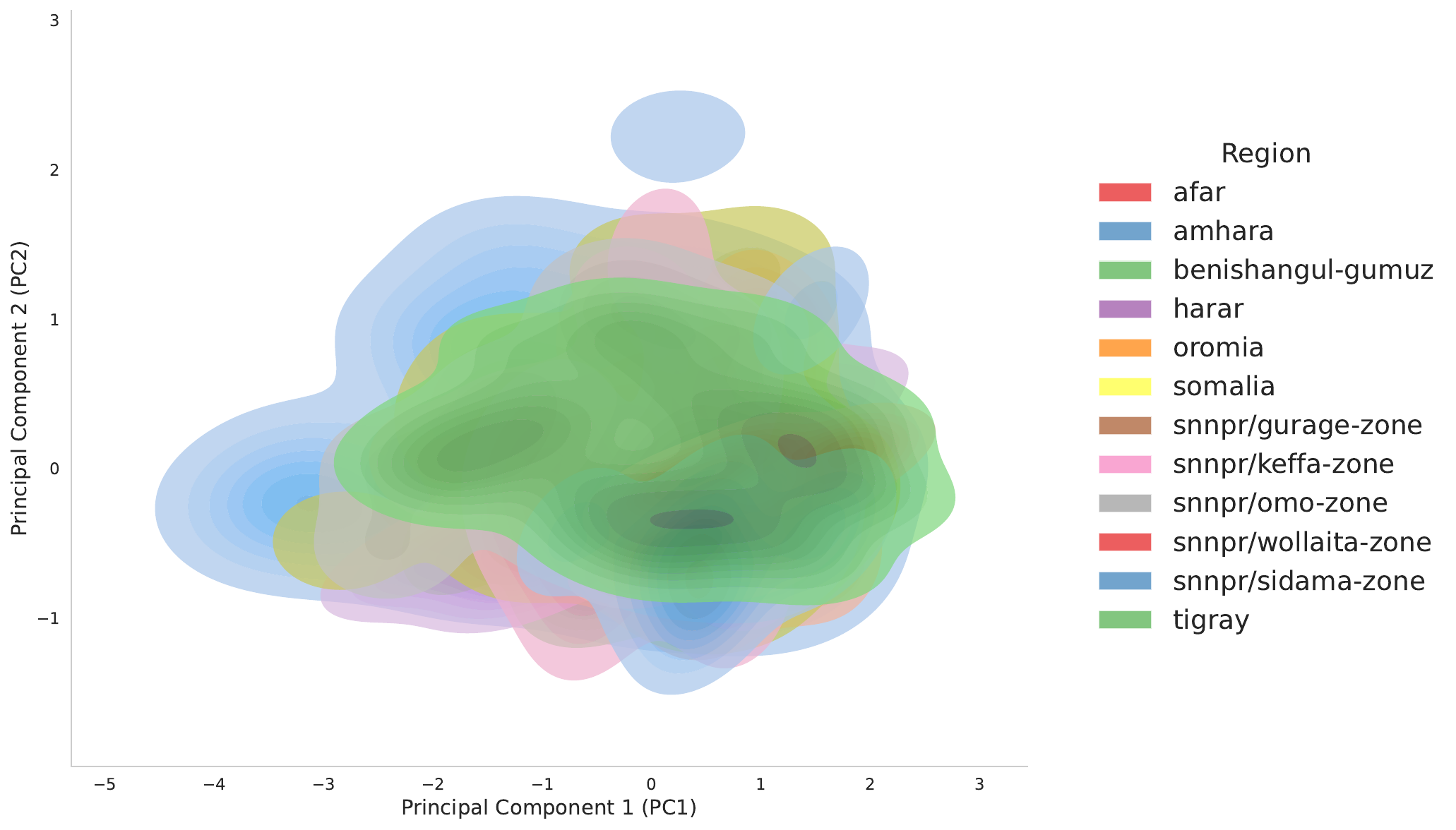}
\caption{
\textbf{Regional semantic density of Amharic stories across regions} in the \textsc{\textit{AmharicStoryQA}} corpus.
Each contour shows a kernel density estimate (KDE) over 2D PCA projections of
multilingual MiniLM sentence embeddings, coloured by region. The plot shows that regional stories largely share a common semantic space, with subtle outer-region contours revealing culturally specific variations within the same language.}
\label{fig:semantic_density_plot}
\end{figure*}
\section{Methodology}

\subsection{Dataset}
The stories used in this dataset are sourced from Ethiopian
folktales\footnote{https://www.ethiopianfolktales.com/am} webpages which was previously included for story generation tasks in low resource LLM work~\citet{azime-etal-2024-walia}.  

As shown in Figure~\ref{fig:AmharistoryQA}, we followed the steps described below, building on prior human-in-the-loop dataset creation approaches~\cite{yuen2025automatic,berkane-etal-2025-llm}.

\begin{figure}
    \centering
    \includegraphics[width=\linewidth]{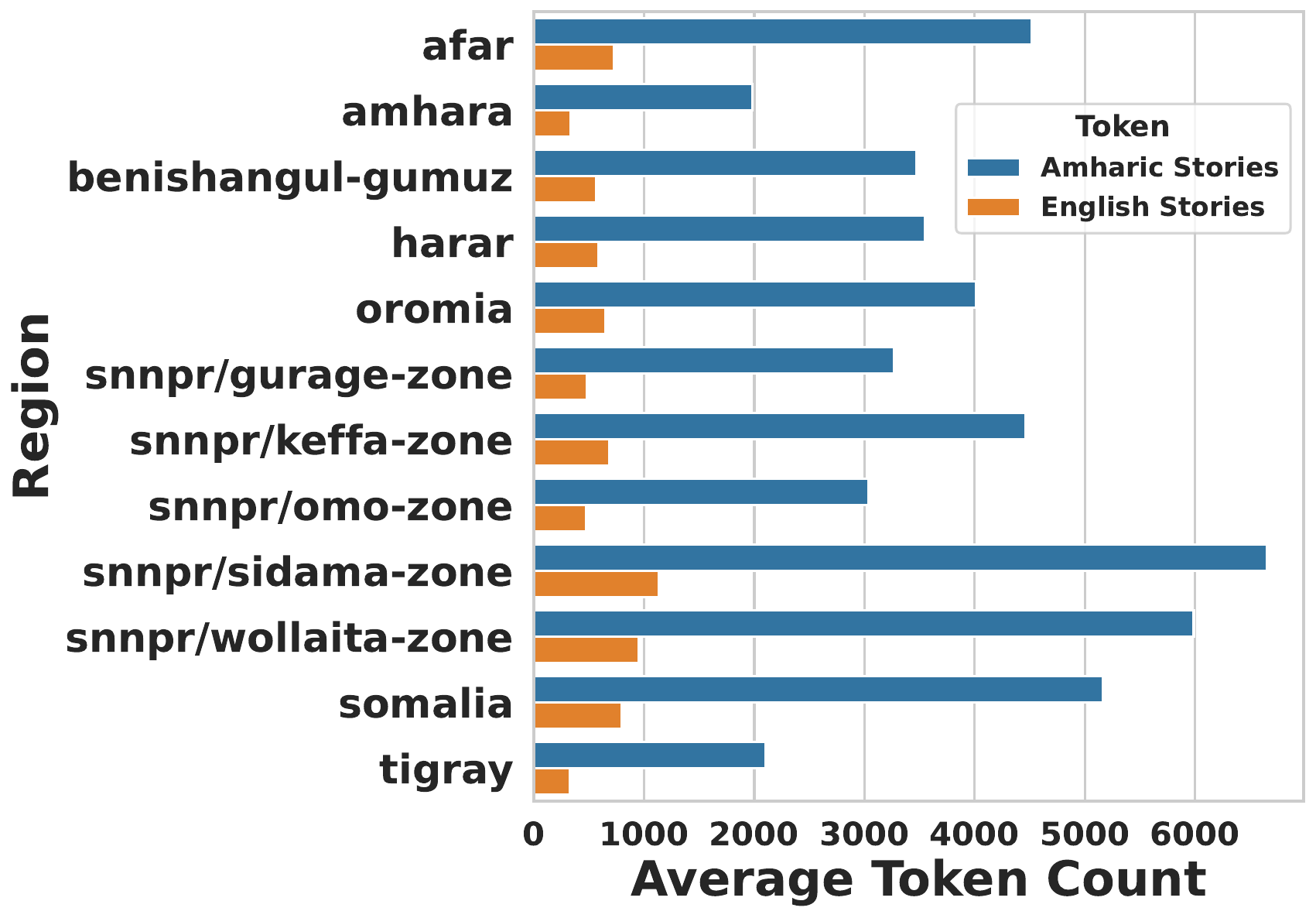}
    \caption{Average Stories Sequence Length by Region}
    \label{fig: token_count}
\end{figure}

We worked with 224 stories collected from nine distinct regions, with the SNNP region further subdivided into multiple zonal classifications, as detailed in Appendix~\ref{sec:ethiopian_regions}. We will release the dataset strictly for research purposes only. Figure~\ref{fig: token_count} illustrates the \llamaeight token counts (sequence lengths) for each story across regions showing coverage of large sequence stories for this work. Average sequence length of stories by region is shown in Table~\ref{fig: token_count}.

\subsubsection{Question Generation}
Given that we have verified English question answer pairs for all Amharic stories collected and presented in Ethiopian folktales webpages, we leveraged \gpt ~to generate batches of five questions per story. This approach enabled us to produce five distinct questions for each story without repeating ideas across them. For each generated question, we also produced one correct answer and three distractor choices: (choice B) a choice biased by world knowledge, (Choice C) an unrelated statement, and (Choice D) a factually false option. We generated 5 questions from each story in parallel to avoid repetitive questions. All content was generated in a single step using structured generation~\footnote{\url{https://platform.openai.com/docs/guides/structured-outputs}}, which helped prevent parsing inconsistencies that can arise when handling multiple separate generations.

\subsubsection{Manual Question Rating}
\label{manual-rating}
After obtaining the English stories, Amharic stories, four distractor choices, and the correct answer, we developed a question-rating guideline to ensure that the generated English questions met standards of correctness and faithfulness, linguistic quality and clarity, and comprehension depth and challenge level. The evaluation was conducted by three human annotators.
\textbf{Correctness and Faithfulness} evaluates whether the question is grounded in the story, whether the correct answer is fully supported by the narrative, and whether the distractors follow the intended structure (plausible, world-knowledge-biased, irrelevant, or factually false).
\textbf{Linguistic Quality and Clarity} assesses whether the question is grammatically correct, fluent, and natural, particularly for Amharic, and whether it is easy to understand and translate.
\textbf{Comprehension Depth and Challenge Level} examines whether the question tests meaningful understanding of the story, including inferential or reasoning-based comprehension rather than simple surface-level recall. We used Gwet’s AC1 to measure agreement between raters beyond chance, as it is more robust to class imbalance and prevalence effects. Exact annotation guideline can be found in Appendix~\ref{annotation-guide}.




\begin{figure}
    \centering
    \includegraphics[width=0.9\linewidth]{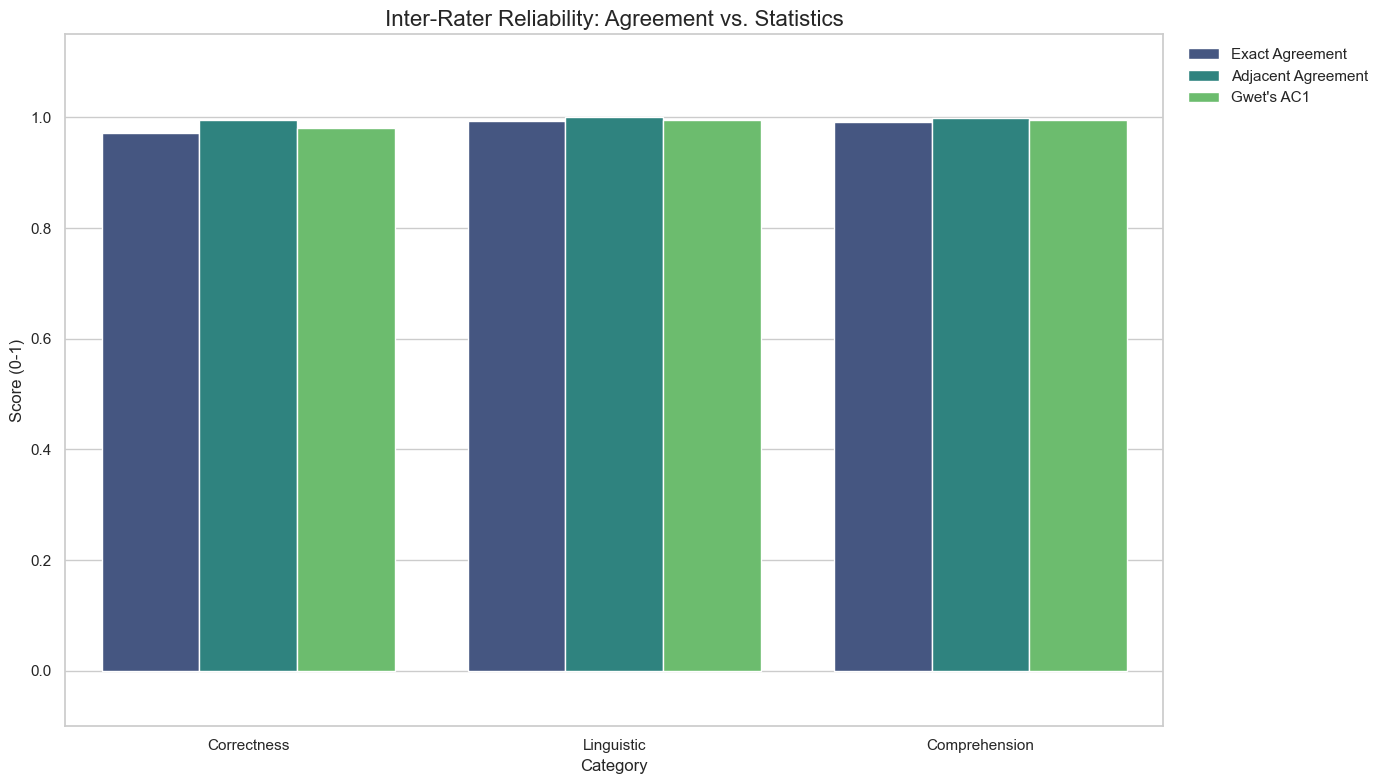}    

    \caption{Inter-Rater Reliability agreement across Correctness, Linguistic, and Comprehension categories assessed with three human evaluators}
    \label{fig:quality}
\end{figure}

\begin{figure*}
\includegraphics[width=0.5\textwidth]{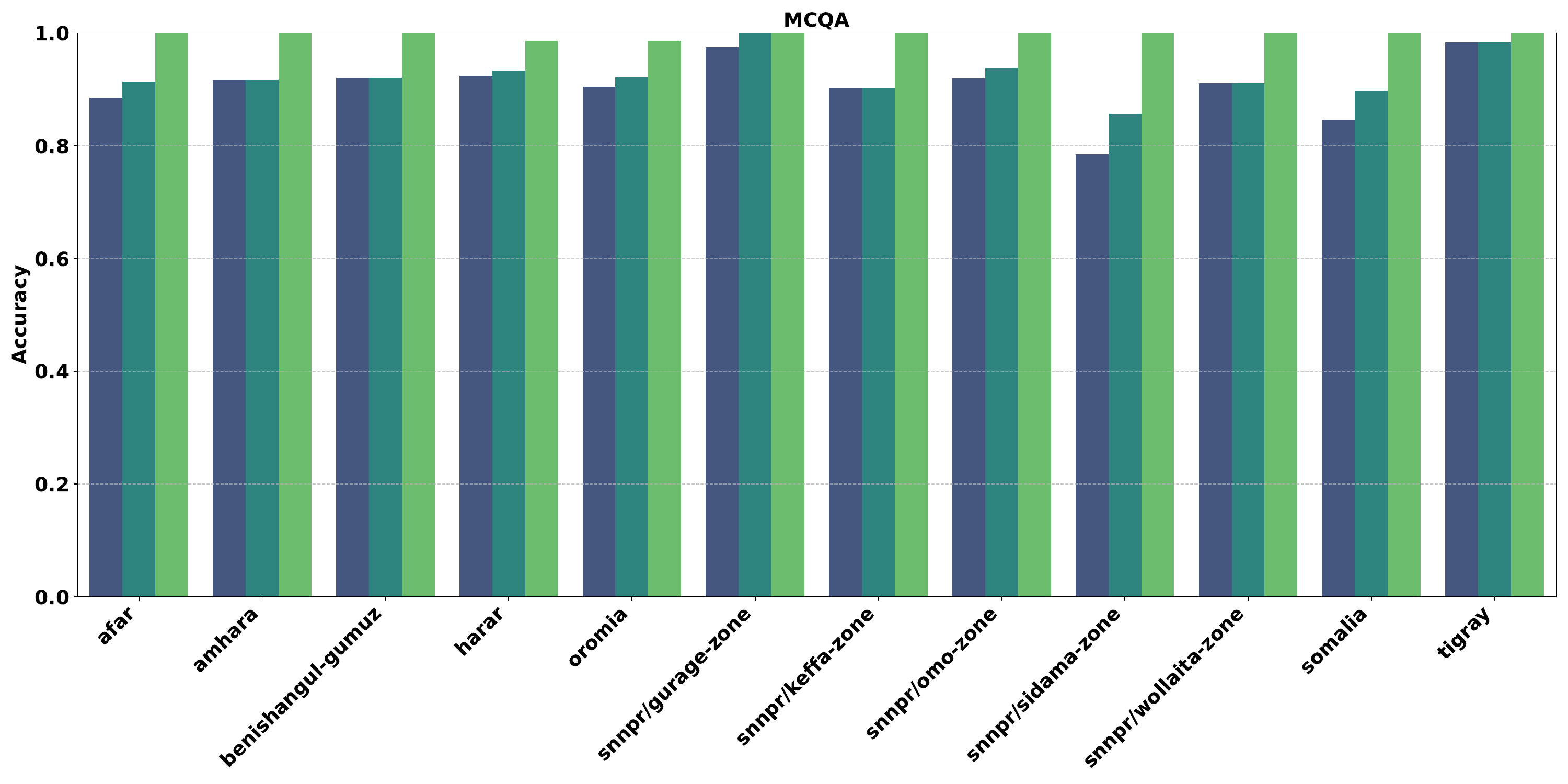}
\includegraphics[width=0.5\textwidth]{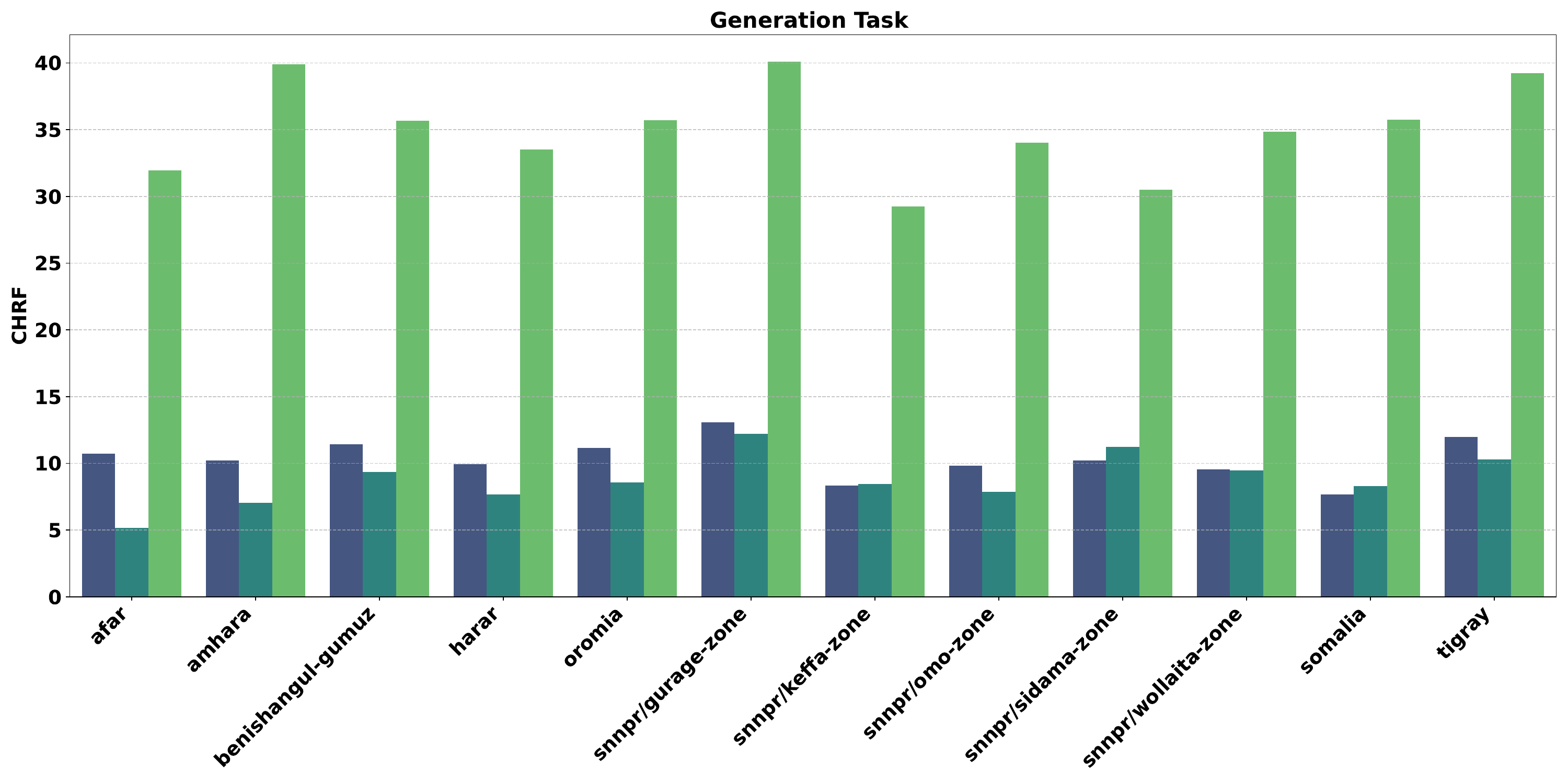}
\includegraphics[width=0.5\textwidth]{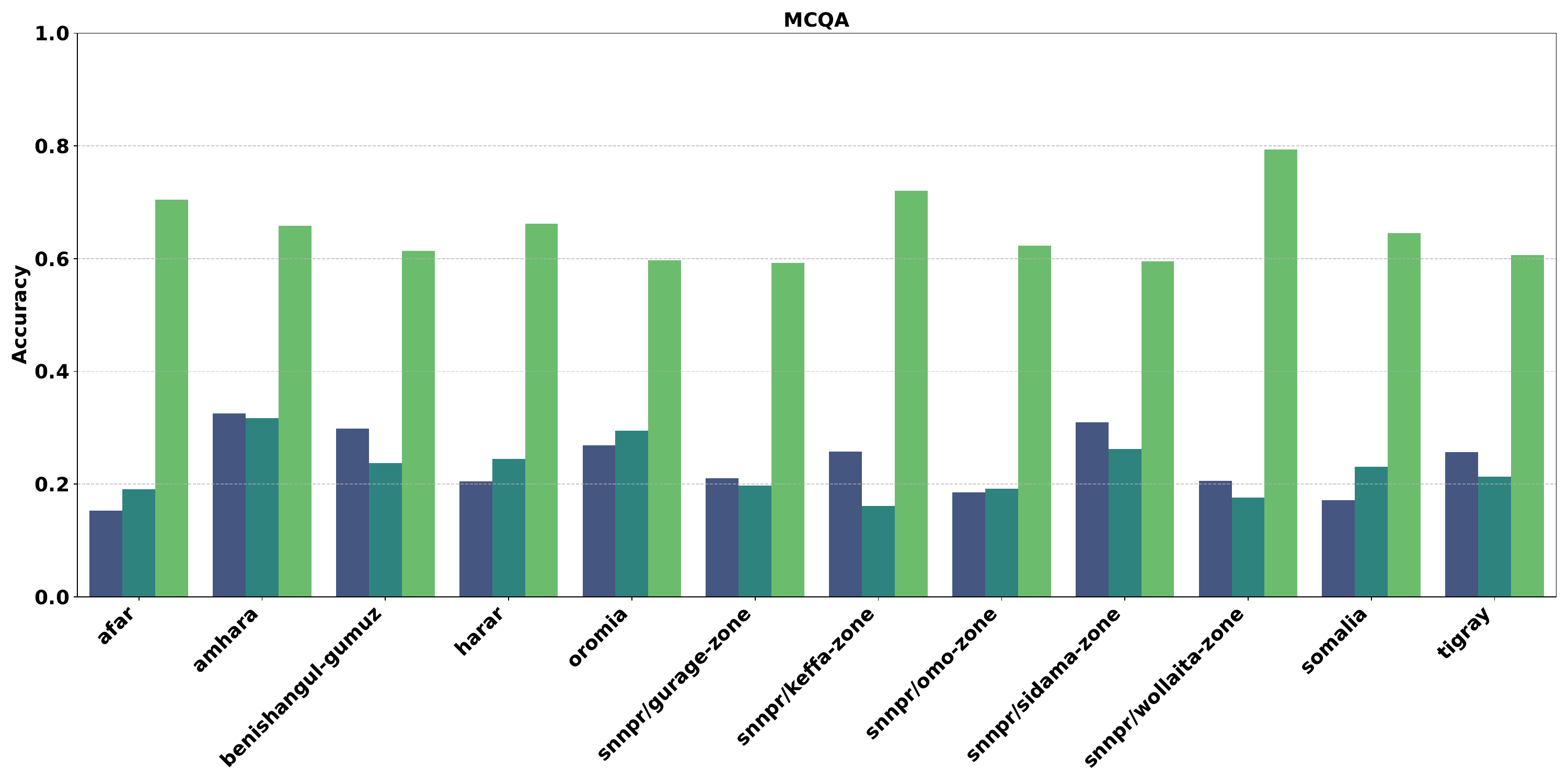}
\includegraphics[width=0.5\textwidth]{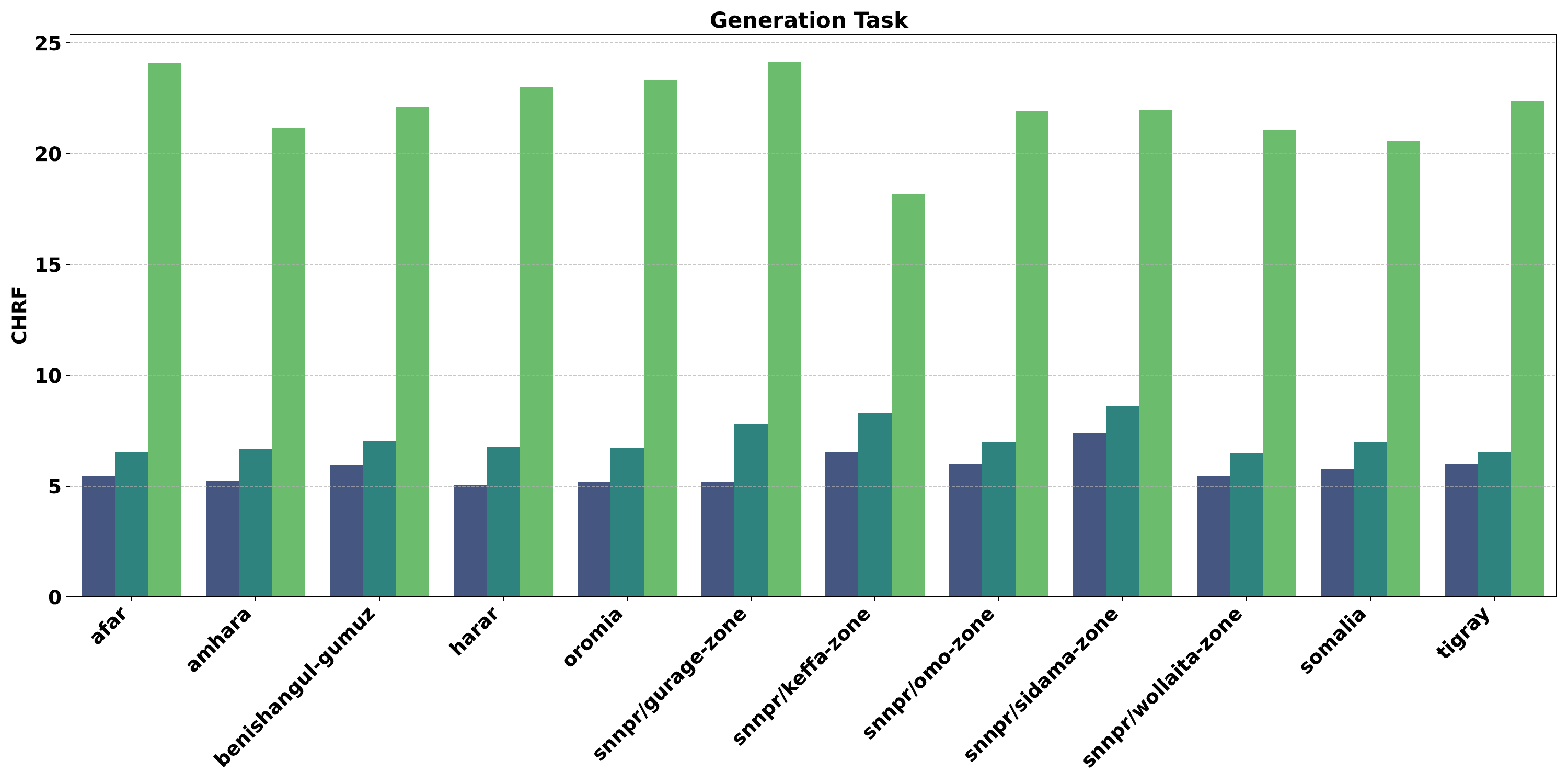}
\end{figure*}
\begin{figure*}
\centering
\includegraphics[width=0.7\textwidth]{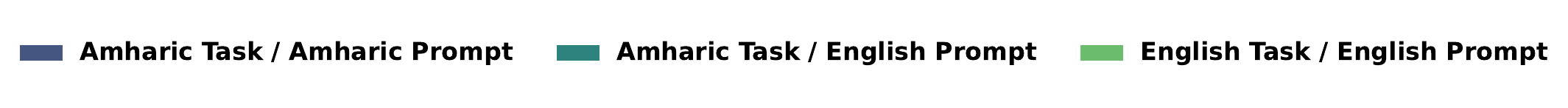}
\caption{\textbf{Regional Performance.} 
Performance of the two differently performing models (top) (\gemmathree) and (\llamaone), showing performance variation across narratives from different regions in our dataset.}
\label{fig:regional result}
\end{figure*}

\subsubsection{Manual Translations}

we translated each question and its answer choices into Amharic using experienced annotators who had previously worked on translation tasks. The translators were familiar with the narrative content, enabling them to preserve meaning, style, and cultural relevance across all generated questions.

\subsubsection{SSA-COMET scoring}
To evaluate machine translation (MT) quality for our low resource African language setting, we adopt both reference-based and reference-free evaluation metrics introduced in \citet{li-etal-2025-ssa}. These metrics allow us to systematically assess the quality of the human-translated Amharic outputs. Following prior work~\cite{alabi-etal-2025-afridoc,azime2025bridging}, we flagged all translations with a score below 0.65 for manual inspection. While we found that many low-scoring cases were in fact false negatives produced by SSA-COMET~\cite{li-etal-2025-ssa}, reflecting known limitations of automatic evaluation in low-resource languages, we nonetheless identified and corrected a small number of genuine translation errors attributable to the translators. This combined automatic-plus-manual verification ensured that our final Amharic translations maintained high linguistic quality and fidelity to the original stories.

\subsubsection{Train-Test Split}

We sorted all story and question–answer pairs by their average SSA-COMET scores and used this ordering to create the train–test split. This approach ensures that translation inconsistencies do not disproportionately influence model evaluation and failures due to translation problems does not result evaluation problem.

After randomizing the answer choices from the structured-generation labels (where the correct answer was always originally assigned to option A) to simplify downstream quality control, we created a balanced 50/50 split for the benchmark. This split was lightly informed by the overall SSA-COMET score distribution to avoid translation-quality imbalance, resulting in 571 training instances and 649 test instances. This include instances where ssa-comet fails to capture translation quality according to human raters.



\begin{figure*}
\includegraphics[width=0.5\textwidth]{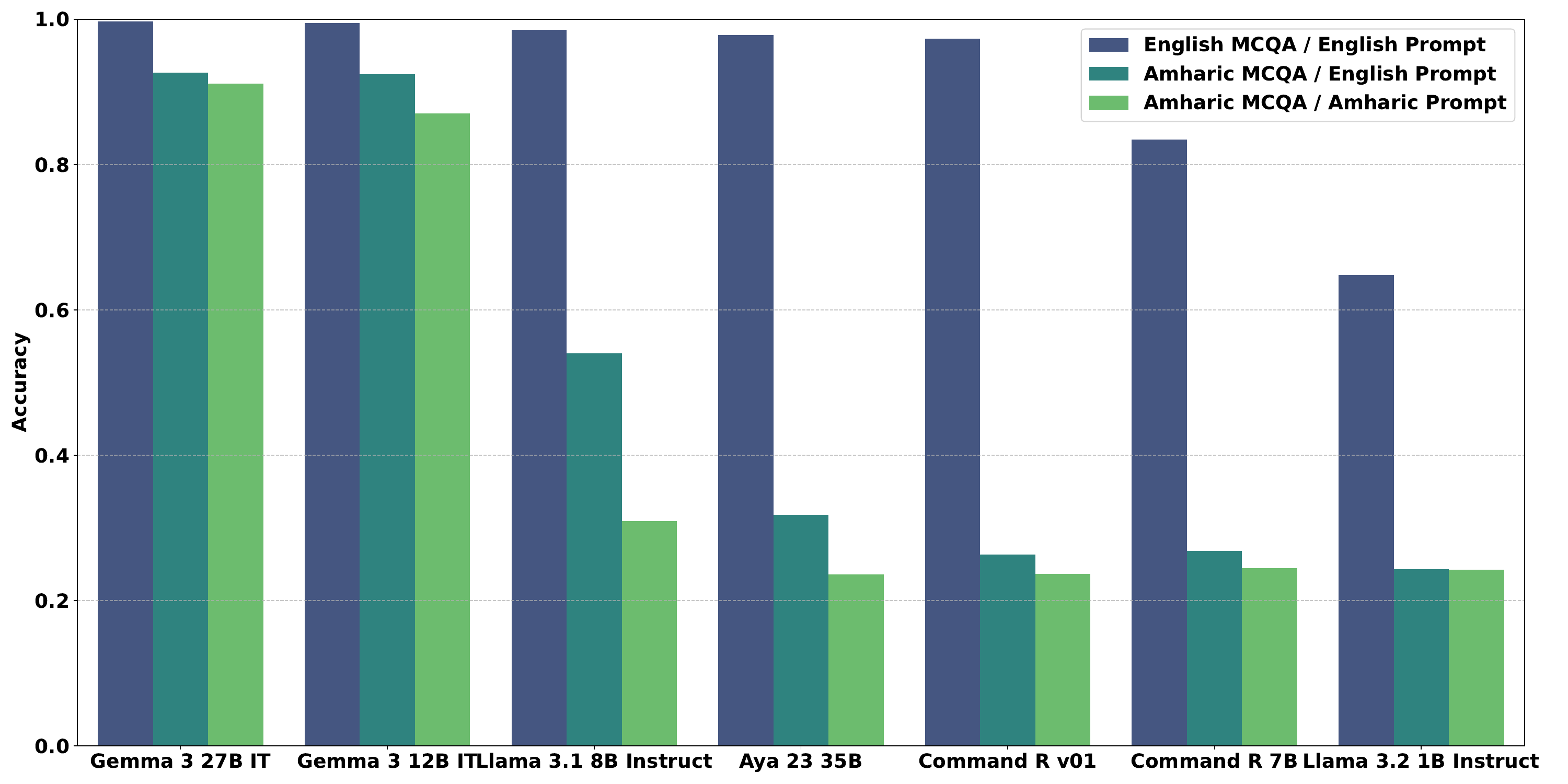}
\includegraphics[width=0.5\textwidth]{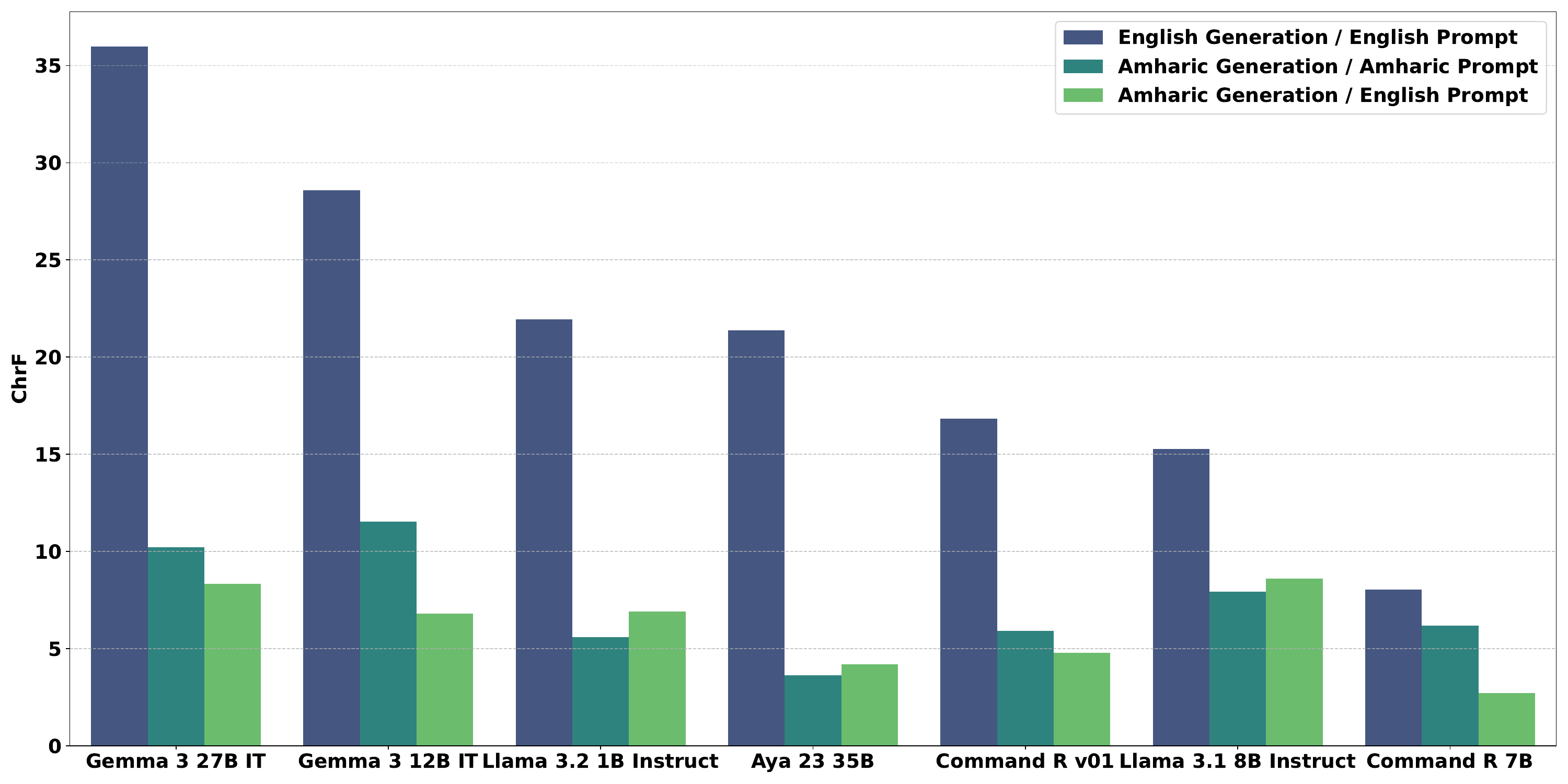}
\caption{\textbf{Results on Amharic vs. English tasks.} 
Left: Accuracy on multiple-choice question answering (MCQA) tasks under different task–prompt language settings. 
Right: Generative performance measured using chrF under the same conditions. }
\label{fig:benchmark-results}
\end{figure*}

\section{Experimental Setup}
\subsection{Evaluation}
We selected \textbf{7} models that support a maximum context length of 128k tokens to align with the long-sequence nature of our tasks.

We conducted all evaluations using EleutherAI’s open-source Language Model Evaluation Harness (\texttt{lm-eval})~\cite{eval-harness}. The framework supports multiple evaluation strategies, including log-likelihood, text generation, and perplexity, with experiments configured and managed via configuration files. For open-source models, we employed log-likelihood and generation-based evaluation for multiple-choice tasks. In multiple-choice  each candidate option is appended to the corresponding question prompt, after which the log-likelihood is computed. Model accuracy is then reported based on the option with the highest log-likelihood score.
\subsection{Finetuning}

To fine-tune LLLMs, we leveraged \textbf{LLaMA-Factory}~\cite{zheng2024llamafactory}, an open-source framework that supports efficient fine-tuning through parameter-efficient methods and scalable training pipelines. We performed LoRA-based supervised fine-tuning using two A100 GPUs on English-only, Amharic-only, and combined multilingual datasets. All datasets were converted to the Alpaca instruction format and will be shared as part of our released resources. Training was conducted for three epochs using a LoRA rank of 8 and an effective batch size of 8 via gradient accumulation. A systematic exploration and optimization of hyperparameters, including LoRA configuration and training dynamics, is left for future work.
\subsection{Task choice}

Previous work has shown that large language models’ performance on multiple-choice question answering (MCQA) tasks is heavily influenced by factors such as option ordering and other structural biases~\cite{loginova-etal-2025-addressing}. These limitations are further amplified in low-resource language settings, as demonstrated by recent analyses in~\cite{azime-etal-2025-proverbeval}. Despite these known issues, MCQA remains one of the most widely used evaluation paradigms due to its ease of benchmarking, reproducibility, and straightforward metric computation. For this reason, we include MCQA tasks in our evaluation while complementing them with generative tasks, allowing for a more balanced assessment across different task formulations and reducing reliance on a single evaluation signal.

\section{Result and Analysis}

\paragraph{Cultural Differences}
Following~\citet{liu-etal-2024-multilingual}, we visualize regional semantic variation using kernel density estimates (KDE) over 2D PCA projections of multilingual MiniLM sentence embeddings. As shown in Figure~\ref{fig:semantic_density_plot}, the substantial overlap among regions suggests that stories written in Amharic largely occupy a shared semantic space, reflecting common narrative structures, themes, and stylistic conventions across Ethiopia. At the same time, the outer contour differences observed for specific regions reveal localized semantic shifts, indicating that certain regions contribute stories with distinctive cultural motifs or domain-specific content. These patterns demonstrate that, although the language is the same, regional storytelling traditions introduce measurable semantic variation.

\begin{figure*}
\includegraphics[width=0.5\textwidth]{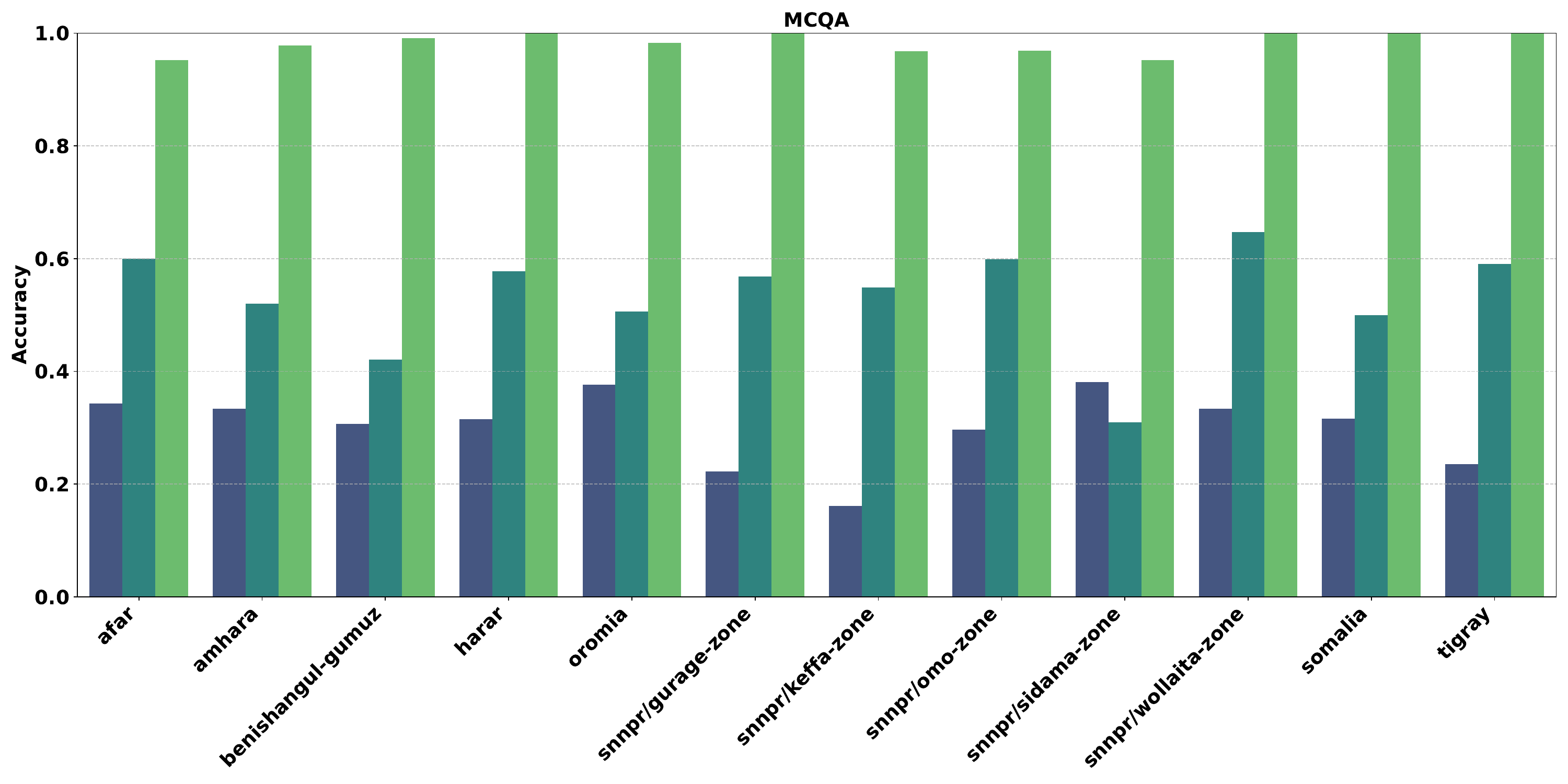}
\includegraphics[width=0.5\textwidth]{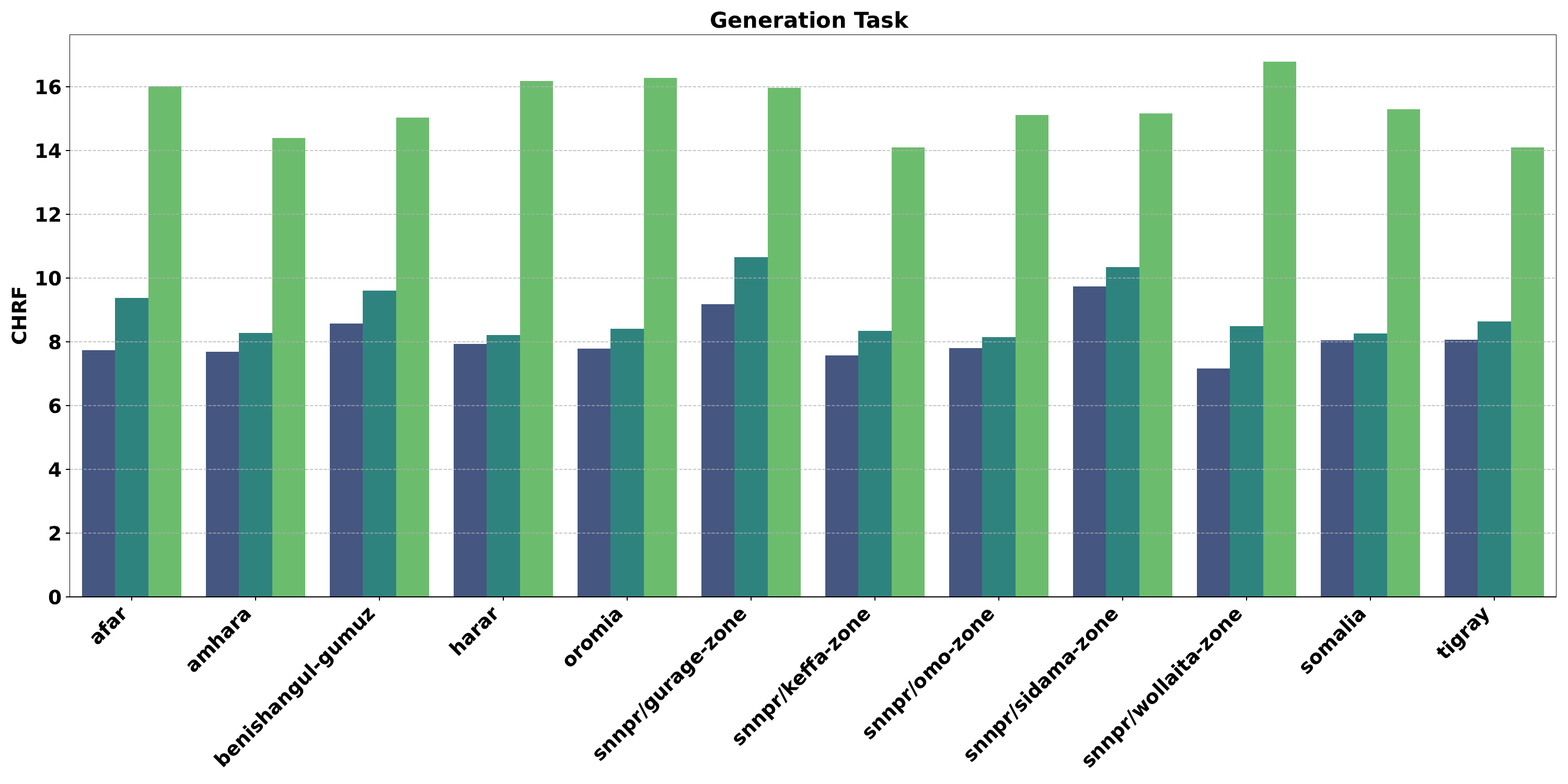}
\includegraphics[width=0.5\textwidth]{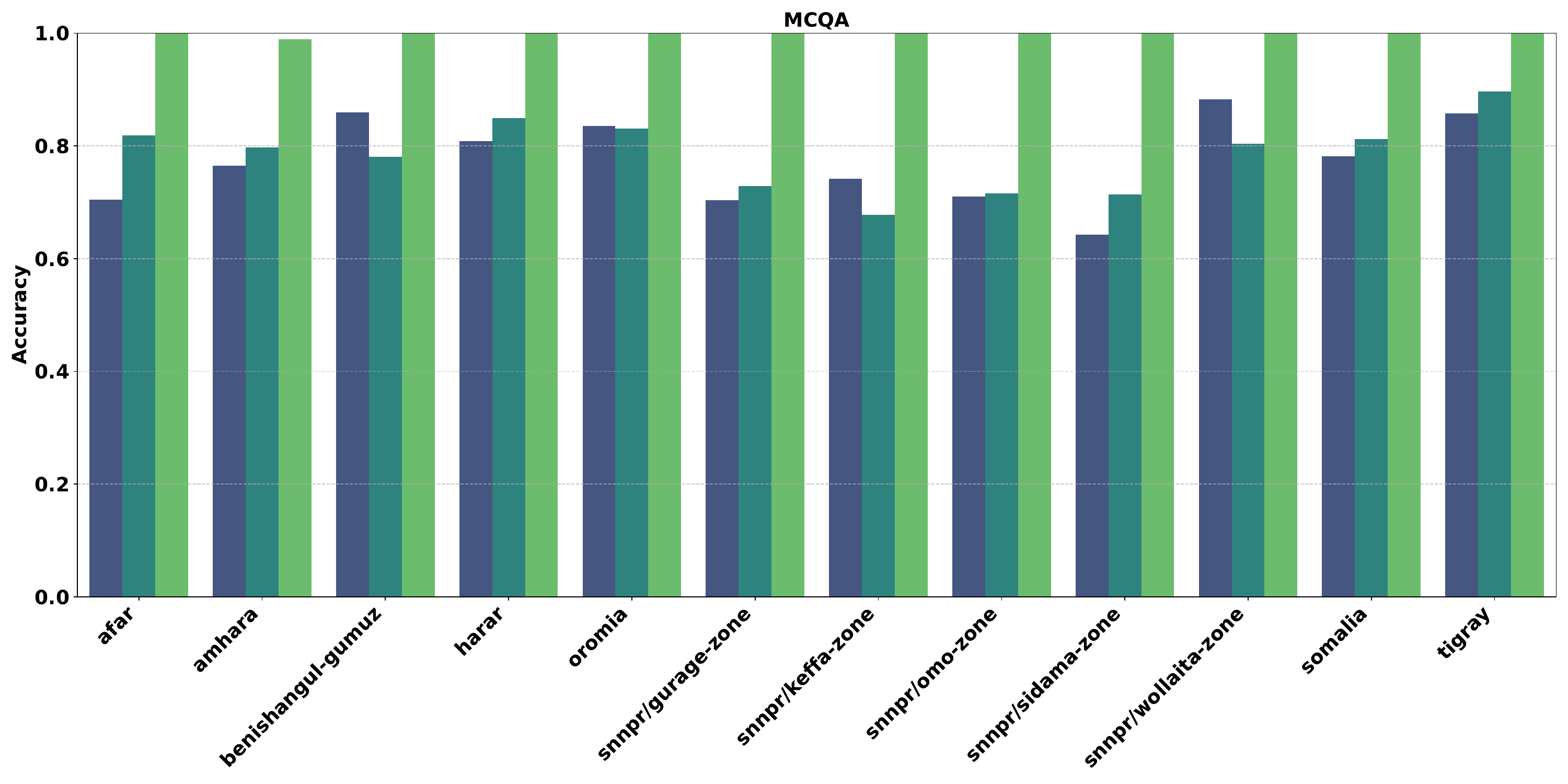}
\includegraphics[width=0.5\textwidth]{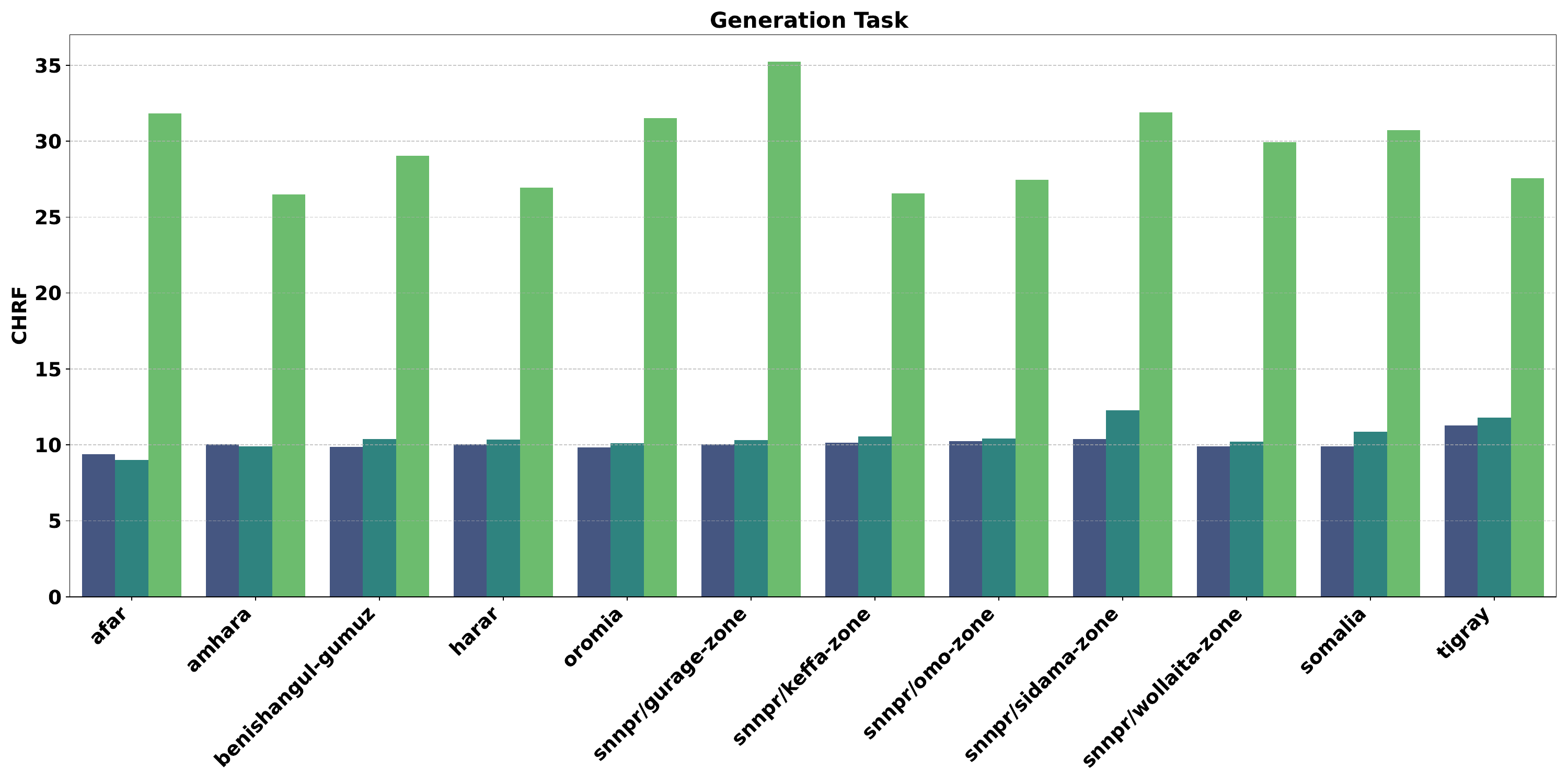}
\end{figure*}
\begin{figure*}
\centering
\includegraphics[width=0.7\textwidth]{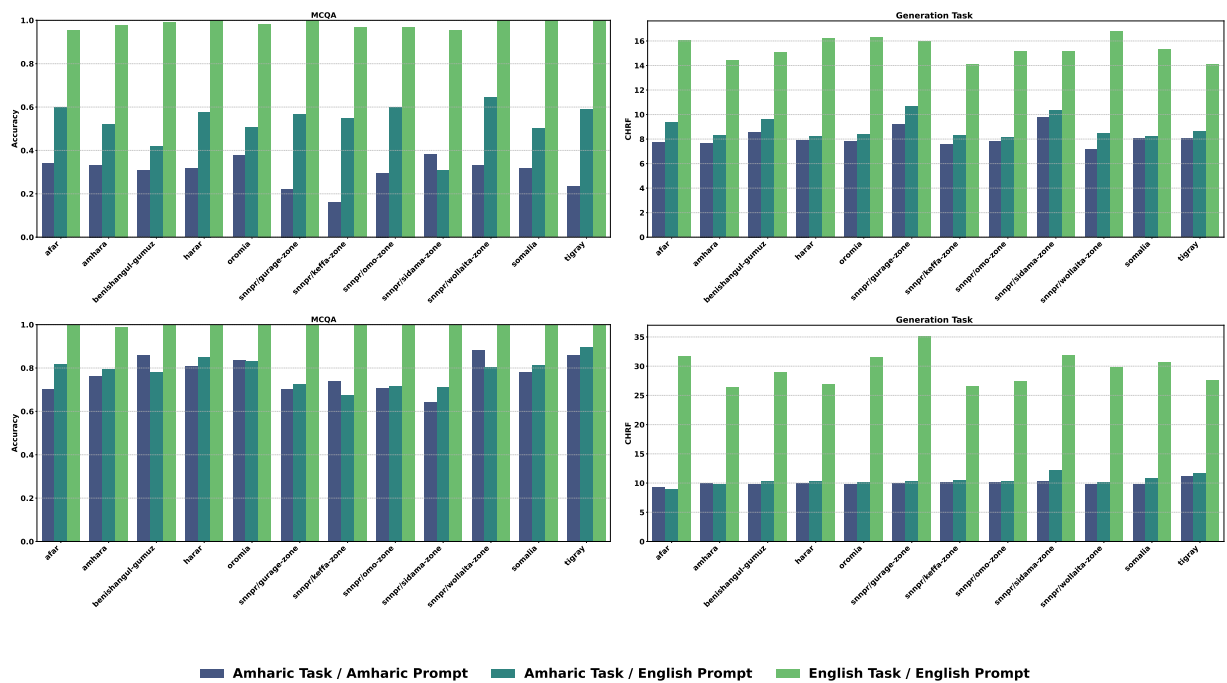}
\caption{\textbf{Effect of SFT on \textit{AmharicStoryQA} across regions.} 
Performance of \llamaeight (top) and its SFT variant trained on \textit{AmharicStoryQA} (\ourall, bottom) on MCQA and generative evaluation.}
\label{fig:regional result sft}
\end{figure*}

\paragraph{(RQ1) How well do Large Language Models comprehend Amharic narratives?}

\paragraph{Multiple-choice question answering (MCQA).}
As shown in the Figure \ref{fig:benchmark-results} (Left), The MCQA results reveal a bimodal pattern in Amharic narrative comprehension. While Gemma models retain high accuracy on Amharic tasks (above 0.80\%), all other models exhibit a sharp performance collapse to the 0.20–0.35 range, despite achieving great accuracy on English tasks. This separation suggests that cross-lingual narrative transfer is not gradual but model-dependent, likely reflecting differences in pretraining data or representational alignment rather than model size alone. Notably, English and Amharic prompting yield nearly identical performance for low-performing models, indicating a hard comprehension bottleneck where prompt language has diminishing returns. Moreover, strong English performance does not reliably predict Amharic comprehension: large multilingual models such as \aya~perform comparably to much smaller models on Amharic MCQA. Together, these findings indicate a structural limitation in Amharic narrative understanding rather than a failure of instruction following. Notably, \llamaeight~exhibits the largest performance gap between English and Amharic prompting, suggesting that its narrative comprehension abilities are disproportionately stronger in English. This gap points to insufficient Amharic narrative exposure during training and motivates RQ3, where we investigate whether increased and targeted Amharic narrative data can mitigate this disparity.

\textbf{Generative comprehension.}
As shown in Figure~\ref{fig:benchmark-results} (right), in contrast to the sharp performance collapse observed in MCQA, generative evaluation using chrF exhibits a more gradual degradation when narratives are presented in Amharic. This pattern indicates a consistent under-performance across all Amharic results, rather than abrupt failure on specific instances. While chrF scores drop substantially from English to Amharic tasks, models retain partial generation capability, suggesting preserved lexical and syntactic knowledge despite limited narrative understanding. Providing English prompts, in most cases, for Amharic narratives yields higher chrF scores than Amharic prompts, with gains of roughly 1–3 points, indicating improved surface-level fluency. However, these improvements do not correspond to higher MCQA accuracy, highlighting a dissociation between generative fluency and true comprehension. 

Across both \textbf{MCQA and generative benchmarks}, models exhibit a consistent and systematic performance degradation when operating on Amharic narratives, preserving relative model rankings and prompt-language effects, which jointly indicate a structural limitation in Amharic narrative understanding rather than benchmark-specific failure.

\begin{table*}[!t]
    \centering
    \scalebox{0.7}{
    \begin{tabular}{l|llll}
    \hline
        \textbf{Model Name (Trained Split)} & \textbf{language} & \textbf{prompt\_language} & \textbf{MCQA (accuracy)} & \textbf{Generation(chrf)} \\ \midrule
        \ourall & English & English & \textbf{0.998} & \textbf{28.931} \\ 
        \oureng & ~ & English & 0.992 & 23.901 \\ 
        \ouramh & ~ & English & 0.988 & 18.023 \\ 
        \llamaeight & ~ & English & 0.986 & 15.283 \\ \midrule
        \ourall & Amharic & English & \textbf{0.802} & \textbf{10.410} \\ 
        \ourall & ~ & Amharic & 0.787 & 10.086 \\ 
        \ouramh & ~ & English & 0.758 & 10.056 \\ 
        \ouramh & ~ & Amharic & 0.748 & 9.758 \\ 
        \oureng & ~ & Amharic & 0.540 & 8.259 \\ 
        \oureng & ~ & English & 0.309 & 7.661 \\ 
        \llamaeight & ~ & Amharic & 0.383 & 7.944 \\ 
        \llamaeight & ~ & English & 0.701 & 8.603 \\ \bottomrule
    \end{tabular}
        }
    \caption{Effect of training \llamaeight on different \textit{AmharicStoryQA} subsets (English-only, Amharic-only, and combined) for narrative understanding.}
    \label{table: improvement}
\end{table*}
\paragraph{(RQ2) How cultural differences influence LLM performance in a single language?}

In Figure~\ref{fig: token_count}, we observe variability in sequence length across stories from different regions, highlighting region specific narrative characteristics beyond the semantic based analysis presented in Figure~\ref{fig:semantic_density_plot}. In this research question, we aim to break down Amharic narrative understanding capabilities and examine how they are influenced by regional differences. Through manual exploration, we observe that stories often focus on region specific topics shaped by local traditions; for example, narratives from the Afar region where pastoralism is prevalent frequently involve camel herding and cattle theft.

To answer this question, Figure~\ref{fig:regional result} examines two models with contrasting characteristics, \gemmathree~and \llamaone, and analyzes their performance across regions. For English narrative understanding, we observe noticeable regional variation in most model–task settings, with differences of up to ±10 chrF points or nearly 20\% accuracy. An exception is the MCQA performance of \gemmathree, which remains relatively stable across regions. In contrast, Amharic narrative understanding exhibits reduced regional variance, likely due to the uniformly low baseline performance of the models. Similarly, differences arising from prompt language choice in Amharic are attenuated in regional evaluations, suggesting that regional effects are overshadowed by broader limitations in Amharic narrative comprehension.

\paragraph{(RQ3) Can we improve LLM performance using the culturally grounded training dataset we introduce, demonstrating the importance of such data for addressing the limitations observed? }

Using both the English and Amharic training splits of the \textit{AmharicStoryQA} dataset, we aim to improve the narrative understanding capabilities of existing LLMs and to demonstrate that our curated dataset can mitigate the narrative comprehension limitations observed in prior evaluations. We select \llamaeight as our primary model due to its comparatively low and variable performance in the benchmarking results shown in Figure~\ref{fig:benchmark-results}. This choice allows us to more clearly assess the impact of our training data. In contrast to models such as the Gemma family, which exhibited stronger baseline performance, \llamaeight provides a suitable testbed for evaluating whether targeted supervision can substantially improve Amharic narrative understanding.

As shown in Table~\ref{table: improvement}, LoRA fine-tuning on the combined and English-only splits leads to clear gains in English narrative comprehension, with the largest improvements observed in generative evaluation, reaching up to +13 chrF. In contrast, MCQA performance shows limited gains, likely due to near-saturated accuracy in English settings.

For Amharic narrative understanding, we observe substantially larger benefits, with MCQA accuracy improving by nearly 40\% compared to the base instruction model. Notably, the performance gap between English and Amharic prompting narrows when training on the combined dataset, indicating reduced prompt-language sensitivity. Moreover, cross-lingual training, using English stories to improve Amharic evaluation and vice versa, consistently improves both MCQA and generative scores, providing evidence of effective cross-lingual narrative transfer.

\paragraph{Fine-tuned models and regional improvements.}
Figure~\ref{fig:regional result sft} shows that fine-tuning on \textit{AmharicStoryQA} both improves MCQA performance and stabilizes results across regions, substantially reducing the regional disparities observed in the base model. This stabilization effect is particularly evident for Amharic narrative understanding, where fine-tuning leads to more consistent performance across diverse regional story distributions.

In contrast, regional differences are more pronounced for English narratives, even after fine-tuning. This suggests that fine-tuning primarily benefits Amharic narrative representations, while also revealing underlying regional characteristics that remain less influential for English benchmarks. As a result, fine-tuned models exhibit clearer regional variation in English tasks, indicating that regional narrative properties are more visible once the model’s overall comprehension improves.

\section{Conclusion and Future Work}

In this work, we investigated multicultural variation within a single language and its impact on long-sequence Amharic story question answering. We introduced \textit{AmharicStoryQA}, a regionally grounded benchmark designed to capture cultural and narrative diversity among Amharic speakers, and used it to systematically evaluate the narrative understanding capabilities of large language models across regions. Our results show that cultural variation within the same language meaningfully affects story comprehension, revealing limitations in current LLMs that are not captured by standard, language-level evaluations. In particular, we find that regional narrative characteristics influence both selection-based and generative understanding, highlighting the need for culturally aware evaluation and training data. For future work, we plan to extend this single-language, multi-culture analysis to a broader multilingual and multicultural setting.


\section*{Limitations}

Due to the scarcity of volunteers and the high level of manual effort required, this study does not extend the analysis to multiple languages. As a result, we are unable to determine whether the observations reported in this work generalize across languages or are specific to Amharic.


\bibliography{custom}
\clearpage
\appendix

\onecolumn
\section*{Appendix}
\section{ Dataset Distribution in Regional}
\begin{figure}[H]
    \centering
    \includegraphics[width=0.6\linewidth]{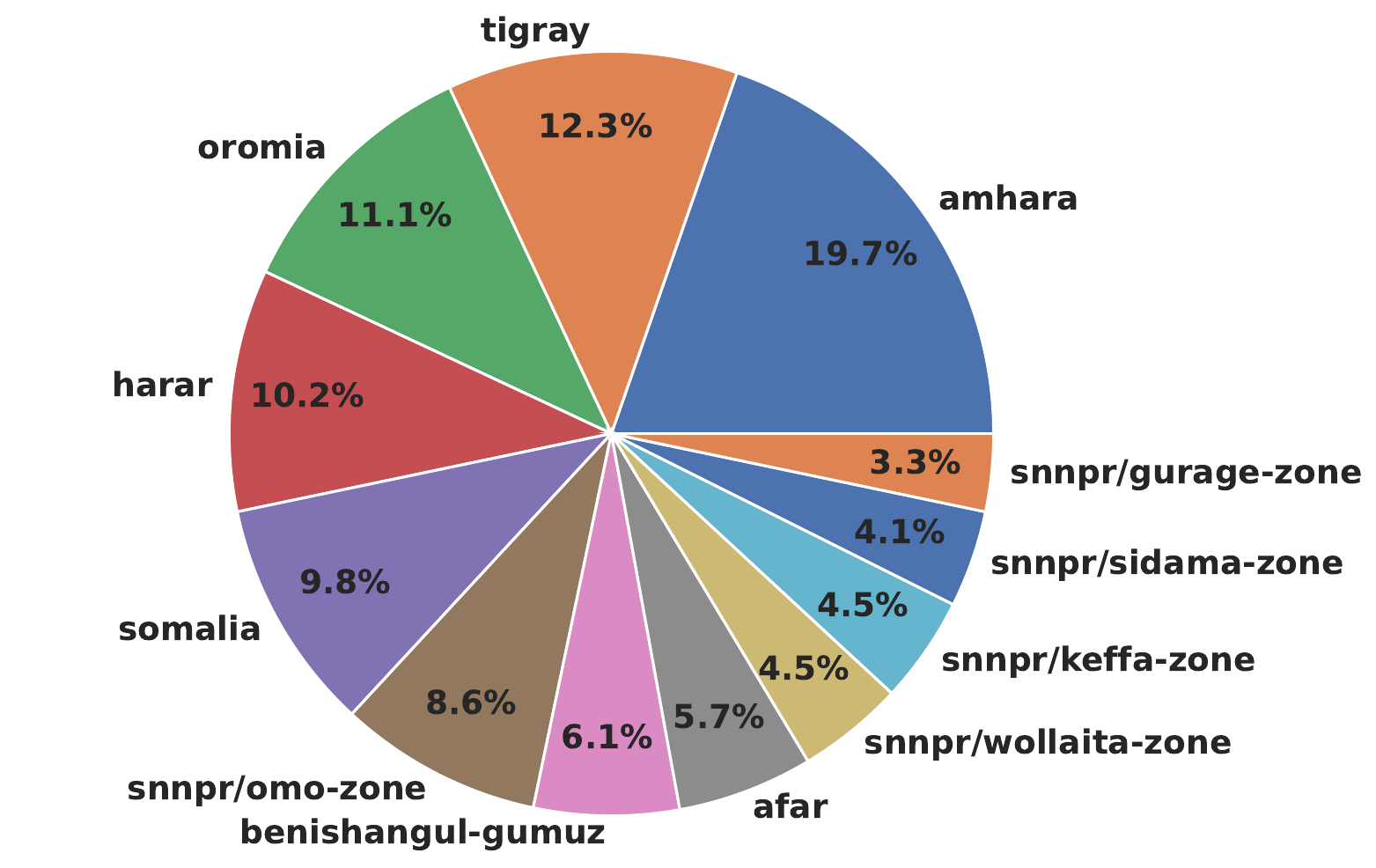}
    \caption{Dataset distribution in Region }
    \label{fig:dataset_distribution}
\end{figure}

Stories depicting cultural practices, daily life, and narratives from the Amhara region constitute approximately 20\% of the total \textit{AmharicStoryQA} dataset. Stories referring specifically to the Tigray and Oromia regions account for 12\% and 11\% of the dataset, respectively. As shown in Figure \ref{fig:dataset_distribution}, the remaining regions of Ethiopia collectively represent more than 50\% of the dataset.

\section{Ethiopian Regions}
\label{sec:ethiopian_regions}
\textbf{Tigray:} is the northernmost regional state, bordering Eritrea to the north. It is the homeland of the Tigrayan people, who speak Tigrinya. The region is historically significant as the center of the ancient Aksumite Empire.

\textbf{Oromia:} is the largest regional state in Ethiopia by both population and land area, stretching across the central, western, and southern parts of the country. It is the homeland of the Oromo people, the largest ethnic group in Ethiopia, and the administrative center is Addis Ababa (Finfinne), which also serves as the national capital.

\textbf{Amhara:} is a regional state located in the northwestern and central highlands of Ethiopia. It is the historic homeland of the Amhara people and is known for its agricultural potential and major tourist attractions like Lalibela and Gondar. The primary language spoken is Amharic.

\textbf{Somali:} is the second-largest region by area, located in the eastern part of Ethiopia bordering Somalia and Djibouti. It is predominantly inhabited by the Somali people who share a common language, culture, and Islamic heritage with Somalis in neighboring countries. The region is largely arid and known for pastoralism.

\textbf{Sidama:} is a regional state in southern Ethiopia formed in 2020 after separating from the former SNNPR. It is the homeland of the Sidama people, who are famous for their high-quality coffee production, a major export for the country. The capital is Hawassa.

\textbf{Afar:} is located in the northeastern part of Ethiopia, home to the Afar people. The region is known for the Danakil Depression, one of the hottest and lowest places on Earth, and is geologically active. The inhabitants are traditionally pastoralists.

\textbf{Benishangul-Gumuz:} is located in western Ethiopia along the border with Sudan. It is home to diverse ethnic groups including the Berta and Gumuz. 

\textbf{Gambela:} is situated in the western tip of Ethiopia, bordering South Sudan. It has a tropical climate and is inhabited primarily by the Nuer and Anuak peoples. The region is rich in wildlife and water resources, including the Baro River.

\textbf{Harari:} is the smallest regional state in Ethiopia, located in the east. It is the homeland of the Harari people and centers around the ancient walled city of Harar (Jugol), a UNESCO World Heritage site and a major center of Islamic culture in the Horn of Africa.

\textbf{South West Ethiopia Peoples' (SWEP):} is a regional state in southwestern Ethiopia formed in 2021. It consists of the Kaffa (the birthplace of coffee), Sheka, Bench Sheko, Dawro, and West Omo zones, representing a diverse mix of ethnic groups and languages.

\textbf{Central Ethiopia:} is a newly formed regional state (established in August 2023) created from the northern part of the former Southern Nations, Nationalities, and Peoples' Region (SNNPR). It is home to the Gurage, Hadiya, and Silte people, among others.

\textbf{South Ethiopia:} is a newly formed regional state (established in August 2023) created from the southern part of the former SNNPR. It includes zones such as Wolayta, Gamo, Gofa, and South Omo, which is famous for its diverse indigenous cultures.

\textbf{Addis Ababa:} is the capital city of Ethiopia and a chartered city (city administration). It serves as the headquarters for the African Union and is a melting pot of all Ethiopian ethnic groups.

\textbf{Dire Dawa:} is the second chartered city in Ethiopia, located in the eastern part of the country. It is a major commercial and industrial hub, historically developed around the Ethiopia-Djibouti Railway.

\section{Avarage Agreement Rating}
\begin{figure}[!ht]
    \centering
    \caption{Average rating of correctness and faithfulness, linguistic quality and clarity across three human evaluators}
    \includegraphics[width=0.5\linewidth]{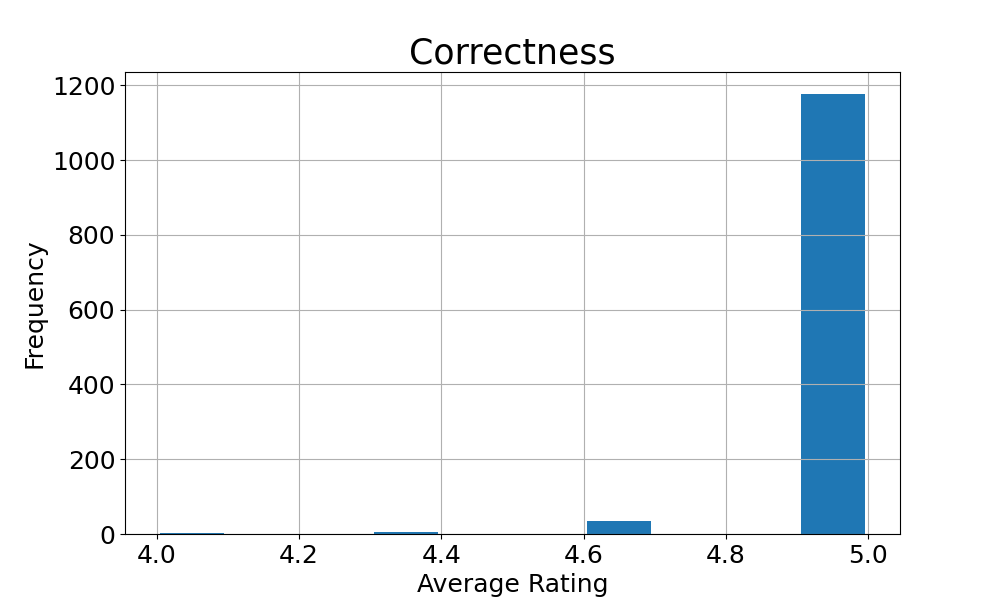}    
    
    \includegraphics[width=0.5\linewidth]{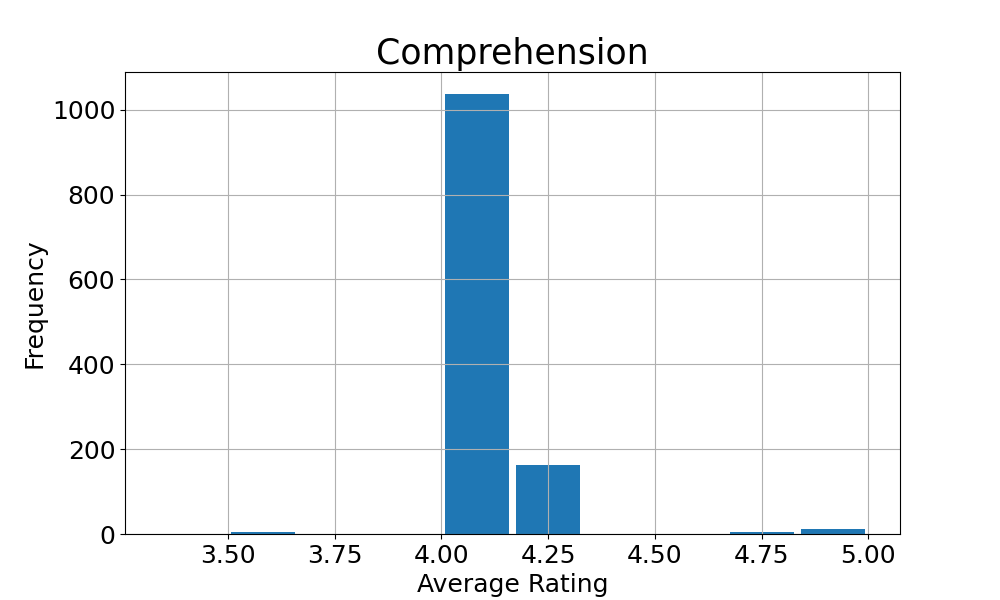}
    
    \includegraphics[width=0.5\linewidth]{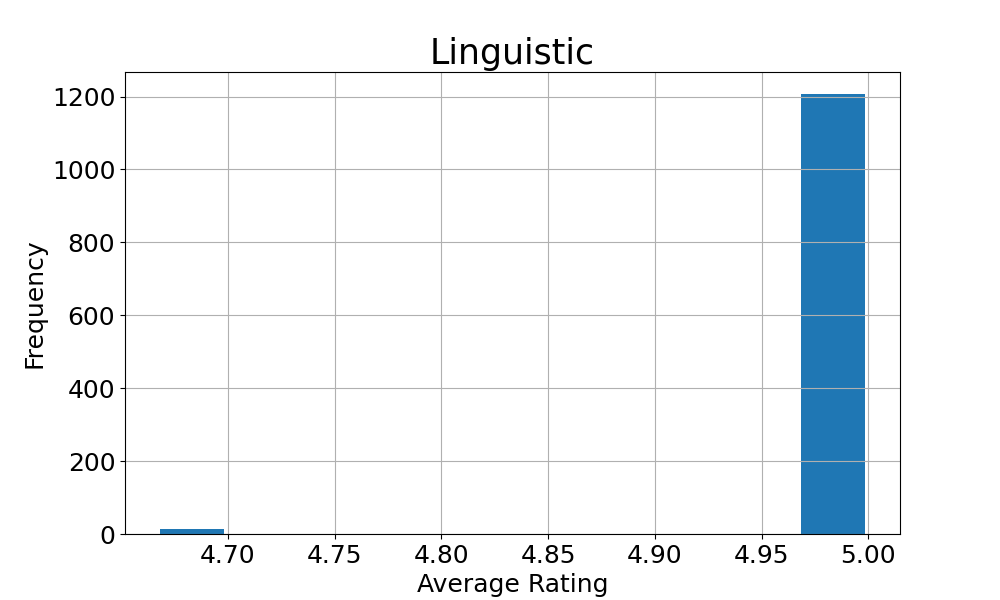}

\end{figure}

\section{Annotation Guidelines}
\label{annotation-guide}
\includepdf[pages=-]{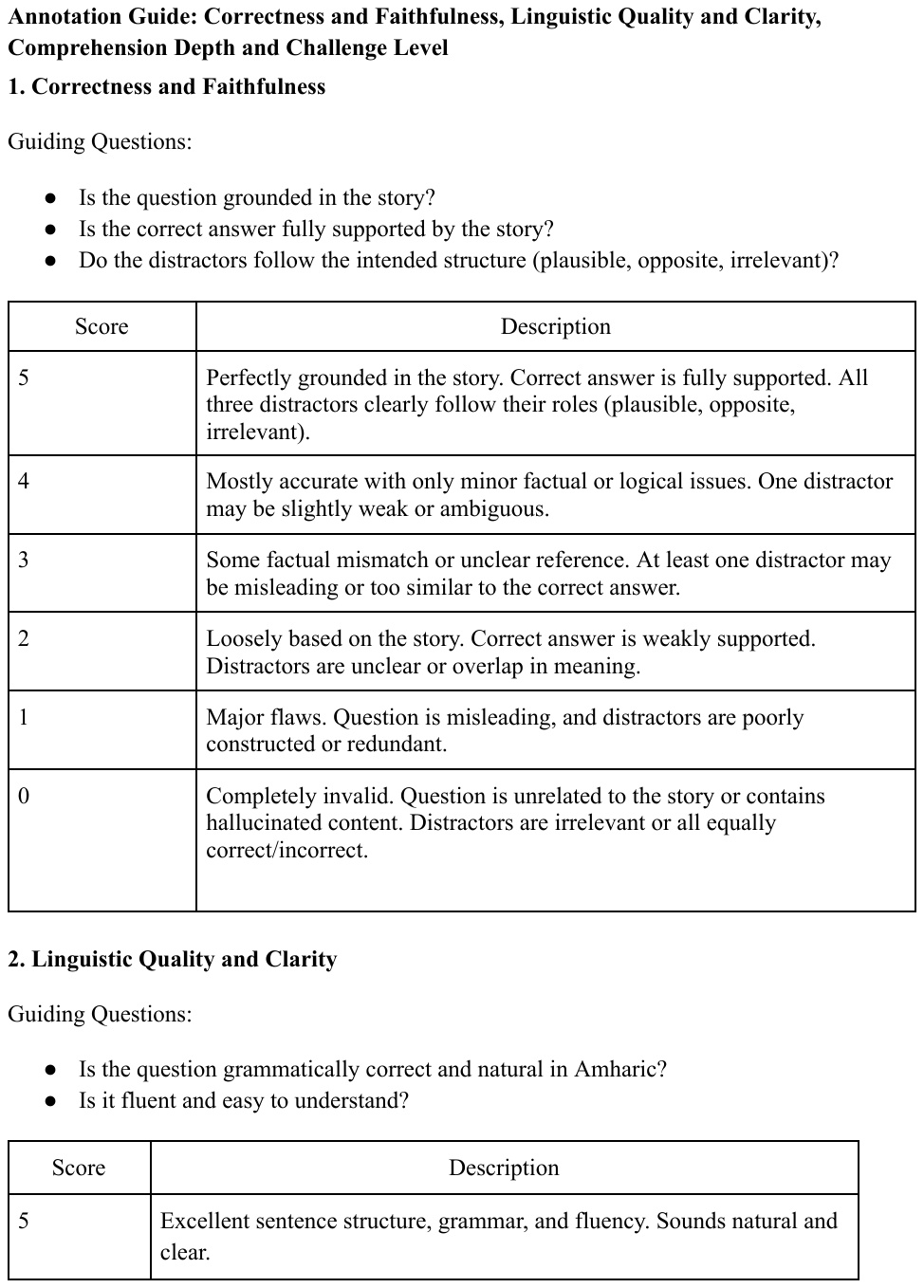}

\section{Training Parameters}

\begin{table}[ht]
\centering

\label{tab:gemma-config}
\begin{tabular}{p{3cm} p{10cm}}
\hline
\textbf{Section} & \textbf{Parameters} \\
\hline
\textbf{Model} & 
\texttt{model\_name\_or\_path:  meta-llama/Meta-Llama-3-8B-Instruct} \newline
\# choose one \\
               & \texttt{trust\_remote\_code: true} \\
\hline
\textbf{Method} & 
\texttt{stage: sft} \newline
\texttt{do\_train: true} \newline
\texttt{finetuning\_type: lora} \newline
\texttt{lora\_rank: 8} \newline
\texttt{lora\_target: all} \\
\hline
\textbf{Dataset} & 
\texttt{dataset: \textit{AmharicStoryQA}\_(split)} \newline
\# choose one or all \newline
\texttt{template: llama3} \newline
\# choose one \newline
\texttt{cutoff\_len: 2048} \newline
\texttt{max\_samples}: none \newline
\texttt{overwrite\_cache: true} \newline
\texttt{preprocessing\_num\_workers: 16} \newline
\texttt{dataloader\_num\_workers: 4} \\
\hline
\textbf{Output} & 
\texttt{output\_dir: \#some directory} \newline
\texttt{logging\_steps: 10} \newline
\texttt{save\_steps: 500} \newline
\texttt{plot\_loss: true} \newline
\texttt{overwrite\_output\_dir: true} \newline
\texttt{save\_only\_model: false} \newline
\texttt{report\_to: none} \\
\hline
\textbf{Train} & 
\texttt{per\_device\_train\_batch\_size: 1} \newline
\texttt{gradient\_accumulation\_steps: 8} \newline
\texttt{learning\_rate: 1.0e-4} \newline
\texttt{num\_train\_epochs: 10.0} \newline
\texttt{lr\_scheduler\_type: cosine} \newline
\texttt{warmup\_ratio: 0.1} \newline
\texttt{bf16: true} \newline
\texttt{ddp\_timeout: 180000000} \newline
\texttt{resume\_from\_checkpoint: null} \\
\hline
\end{tabular}
\caption{Configuration for fine-tuning \texttt{gemma-2-9b-it} and \texttt{Meta-Llama-3-8B-Instruct}}
\end{table}

\section{Evaluation Prompt}
\label{prompt}


\section*{Supplementary material}

**Prompt 1 -- English:**

Read the following story carefully and answer the question that follows.

Story:
"{story}"

Question:
"{question}"

Choices:
A) {choice1}
B) {choice2}
C) {choice3}
D) {choice4}

Your answer (A, B, C, or D):


**Prompt 1 -- Amharic:**

የሚከተለውን ታሪክ በጥንቃቄ ያንብቡ እና ከዚያ በኋላ የሚከተለውን ጥያቄ ይመልሱ።

ታሪክ:
"{story}"

ጥያቄ:
"{question}"

አማራጮች:
A) {choice1}
B) {choice2}
C) {choice3}
D) {choice4}

መልስዎ (A, B, C, ወይም D):


**Prompt 2 -- English:**

Read the story below carefully, paying attention to the details of what happens, and then answer the question that tests your understanding of it.

Story:

"{story}"

Question:
"{question}"

Choices:
A) {choice1}
B) {choice2}
C) {choice3}
D) {choice4}

Your answer (A, B, C, or D):

**Prompt 2 -- Amharic:**

ከታች ያለውን ታሪክ በጥንቃቄ ያንብቡ፣ የተከሰቱን ነገሮች ዝርዝሮች በትክክል ይጠብቁ፣ ከዚያም የገባቃትን ለመፈተሽ የሚከተለውን ጥያቄ ይመልሱ።

ታሪክ:
"{story}"

ጥያቄ:
"{question}"

አማራጮች:
A) {choice1}
B) {choice2}
C) {choice3}
D) {choice4}

መልስዎ (A, B, C, ወይም D):

**Prompt 3 -- English:**

Carefully read the following passage to understand the events, emotions, and main ideas it conveys, then choose the most accurate answer to the question that follows.

Story:
"{story}"

Question:

"{question}"

Choices:
A) {choice1}
B) {choice2}
C) {choice3}
D) {choice4}

Your answer (A, B, C, or D):


**Prompt 3 -- Amharic:**
የሚከተለውን ጽሑፍ በጥንቃቄ ያንብቡ እና ውስጡ የተገለጹትን ክስተቶች፣ ስሜቶች እና ዋና ሀሳቦች በትክክል ለመረዳት ይሞክሩ፣ ከዚያም የሚከተለውን ጥያቄ ለመመለስ ትክክለኛውን አማራጭ ይምረጡ።

ታሪክ:
"{story}"

ጥያቄ:
"{question}"

አማራጮች:
A) {choice1}
B) {choice2}
C) {choice3}
D) {choice4}

መልስዎ (A, B, C, ወይም D):

## F. Question Generation Prompt

You are an expert educational content creator and an analytical text comprehension specialist. Your task is to generate challenging multiple-choice questions from the provided stories. These questions are specifically designed to assess and diagnose language model reading comprehension.

Question Design Guidelines

For each question, follow these rules:

- Focus on important aspects of the story, such as:
    - Characters
    - Events
    - Causes
    - Motivations
    - Outcomes
    - Ensure the question cannot be answered without understanding key parts of the story.
    - Avoid vague or surface-level questions (e.g., "What color was the sky?").

Answer Choice Requirements

Each question must have exactly four answer choices, labeled A to D.

Each option must serve a specific and mutually exclusive diagnostic role, which is essential for evaluating LLM comprehension and reasoning behavior.

**Answer Choice Roles**

A. Correct Answer (Directly Verifiable)

- Role: The *only correct answer* to the question.
- Criteria:
    - Explicitly stated or directly derivable from the story.
    - Requires no external knowledge or unstated assumptions.

- Fully supported by facts in the passage.

- Avoid:
  - Ambiguity
  - Inferential leaps
  - Reliance on common sense beyond basic linguistic understanding.

B. Common-Sense Distractor (External Inference)

- Role: Tests whether the model prioritizes general world knowledge over textual evidence.
- Criteria:
  - Sounds highly plausible and consistent with real-world expectations.
  - Not mentioned, supported, or logically inferable from the story.
  - Introduces information that is external to the text.

- Avoid:
  Using keywords or minor details from the story that could make the option seem textually grounded.

C. Unrelated Hallucination (Random Statement)

- Role: Tests the model's ability to identify and reject irrelevant information.
  Criteria:
  - Completely unrelated to the story's content, characters, or themes.
  - Should not be plausible within the story context.
  - Contains no elements drawn from the text.

- Avoid:
  - Any connection, direct or indirect, to the story.

D. Factual Contradiction (Directly False by Story)

- Role: Tests the model's ability to detect and reject information that contradicts the passage.

- Criteria:
  - Provably and unambiguously false given the story.
  - Directly negates or contradicts an explicitly stated fact.

- Avoid:
  - Being obviously false (it should still be a plausible trap).
  - Inference-based falsehoods (the contradiction must be clear from the text itself, not from external knowledge).

Input

Story: {story}

Output Instructions

- Generate {num_questions} multiple-choice questions.
- Each question must follow all guidelines above.
- Return only a JSON list as the final output.



\end{document}